\documentclass[11pt]{article}

\usepackage[preprint]{acl}

\usepackage{times}
\usepackage{latexsym}

\usepackage[T1]{fontenc}

\usepackage[utf8]{inputenc}

\usepackage{microtype}

\usepackage{inconsolata}

\usepackage{graphicx}
\usepackage{xcolor}
\usepackage{booktabs}
\usepackage{pifont}
\usepackage{tabularx}
\usepackage[normalem]{ulem}
\usepackage{tikz}
\usetikzlibrary{arrows.meta}
\usepackage{placeins}
\usepackage{stfloats}
\usepackage{afterpage}
\usepackage{listings}
\lstdefinestyle{paperprompt}{
    basicstyle=\fontsize{8}{9.2}\selectfont\ttfamily,
    columns=fullflexible,
    keepspaces=true,
    breaklines=true,
    breakatwhitespace=false,
    breakautoindent=false,
    breakindent=0pt,
    showstringspaces=false,
    frame=single,
    frameround=tttt,
    framerule=0.35pt,
    framesep=5pt,
    rulecolor=\color{black!25},
    backgroundcolor=\color{black!2},
    xleftmargin=5.35pt,
    xrightmargin=5.35pt,
    abovecaptionskip=0pt,
    belowcaptionskip=0pt,
    aboveskip=6pt,
    belowskip=8pt
}

\title{Can LLMs Uncover Hidden Physical Laws through Free Exploration?}

\author{
  \textbf{Zhen Yang\textsuperscript{1}},
  \textbf{Wen Wang\textsuperscript{3}},
  \textbf{Xander Xu\textsuperscript{2}\thanks{Co-corresponding authors.}},
  \textbf{Ying-Cong Chen\textsuperscript{1,4}\footnotemark[\value{footnote}]}
\\
  \textsuperscript{1}HKUST(GZ),
  \textsuperscript{2}Alibaba Group,
  \textsuperscript{3}ZJU,
  \textsuperscript{4}HKUST
\\
  \small{
    \href{mailto:zheny.cs@gmail.com}{zheny.cs@gmail.com},
    \href{mailto:yingcongchen@ust.hk}{yingcongchen@ust.hk}
  }
}

\begin{document}
\maketitle

\begin{abstract}
Scientific discovery often begins with an unexpected observation that prompts a new research question.
In a world with altered physical laws, can large language models (LLMs) notice unexpected behavior, such as heavier objects falling more slowly, and discover the law through experiments without being told what to investigate?
To evaluate this ability, we introduce LongChang, an automated framework that creates simulated worlds with altered physics, where one or more laws differ from ordinary physics.
The evaluated LLMs act as explorers, manipulating objects and building tools to investigate their own questions without research tasks or hints about these laws.
Separate LLM agents generate and review scenes, review tool designs, and evaluate reports for correctness and experimental support using the altered physical laws and exploration records.
These records guide scene evolution for further exploration, allowing the full cycle to run without human intervention.
Among 12 LLMs, GPT-6-Astra achieves a discovery rate of 16.7\%, while the other 11 score 0\%.
The exploration records reveal uninvestigated anomalies and only partial identification of the altered physical laws.
Our \href{https://zhenyangcs.github.io/LongChang-website/}{\textcolor{blue}{project page}} and \href{https://github.com/EnVision-Research/LongChang}{\textcolor{blue}{code}} are available.
\end{abstract}

\begin{table*}[t]
    \centering
    \begingroup
    \fontsize{9}{10.5}\selectfont
    \definecolor{sceneink}{HTML}{213949}
    \definecolor{sceneaccent}{HTML}{365F76}
    \definecolor{sceneborder}{HTML}{AFC9D9}
    \definecolor{scenebackground}{HTML}{EAF3F9}
    \definecolor{scenebadge}{HTML}{D5E8F4}
    \definecolor{scenebadgeink}{HTML}{3D6D9B}
    \definecolor{scenecontrol}{HTML}{E5EBDD}
    \definecolor{scenecontroledge}{HTML}{67765C}
\definecolor{scenered}{HTML}{B6525B}
\definecolor{sceneblue}{HTML}{4D7F9E}
\definecolor{scenegreen}{HTML}{4D8569}
\definecolor{sceneyellow}{HTML}{C2A046}
\definecolor{sceneguide}{HTML}{8FA5B4}
\definecolor{sceneneutral}{HTML}{6B737A}
\newcommand{\scenesymbol}[1]{%
    \begin{tikzpicture}[baseline=0pt,x=1pt,y=1pt,line width=0.8pt,
        line cap=round,line join=round,rounded corners=0pt,draw=sceneaccent,fill=none,
        mechanism arrow/.style={-{Latex[length=3.4pt,width=2.6pt]}},
        mechanism guide/.style={draw=sceneguide,line width=0.45pt},
        every node/.style={font=\fontsize{9}{10.5}\selectfont,text=sceneink}]
        \path[use as bounding box] (0,0) rectangle (76,18.5);
        \begin{pgfinterruptboundingbox}
        \ifcase\numexpr#1\relax
        \or 
            \node[text=scenecontroledge] at (38,9) {No law altered};
        \or 
            \fill[scenered] (15,9.25) circle (2.5);
            \fill[scenered] (61,9.25) circle (2.5);
            \draw[mechanism arrow,scenered] (20,9.25) -- (32,9.25);
            \draw[mechanism arrow,scenered] (56,9.25) -- (44,9.25);
        \or 
            \node at (38,9) {heavier $\rightarrow$ slower};
        \or 
            \draw[mechanism guide] (14,17) -- (14,2) -- (65,2);
            \node at (6,11) {$g$};
            \node at (71,5) {$t$};
            \draw[mechanism guide,densely dashed] (14,10) -- (65,10);
            \draw (14,10) .. controls (20,17) and (27,17) .. (33,10)
                .. controls (39,3) and (46,3) .. (52,10)
                .. controls (56,14) and (61,16) .. (65,14);
        \or 
            \draw[mechanism guide] (15,17.5) -- (15,7) -- (65,7);
            \node at (5.5,13) {drag};
            \node at (64,2.7) {speed};
            \draw[mechanism guide,densely dashed] (36,7) -- (36,17);
            \draw (15,7) -- (36,7) -- (65,17);
            \fill[sceneaccent] (36,7) circle (1.2);
            \node at (36,2.7) {2.5};
        \or 
            \node at (45,17) {top view};
            \node at (21,17) {$v$};
            \fill[sceneneutral] (10,12) circle (2.4);
            \draw[mechanism guide,densely dashed] (14,12) -- (70,12);
            \draw[mechanism arrow] (14,12) -- (26,12);
            \draw[mechanism arrow] (14,12) .. controls (44,12) and (59,11) .. (64,2);
            \draw[mechanism arrow] (10,8.8) -- (10,1);
            \draw[draw=sceneaccent,line width=0.6pt] (15,12) -- (15,7) -- (10,7);
            \node at (4.5,4.5) {$F$};
        \or 
            \draw[scenered,line width=1.8pt] (8,1.2) -- (32,1.2);
            \draw[sceneblue,line width=1.8pt] (44,1.2) -- (69,1.2);
            \draw[mechanism guide,densely dotted] (9,16) -- (68,16);
            \fill[sceneneutral] (11,16) circle (1.6);
            \fill[sceneneutral] (47,16) circle (1.6);
            \draw[sceneneutral] (11,16) .. controls (14,14) and (18,7) .. (19,2.8);
            \draw[sceneneutral,mechanism arrow] (19,2.8) .. controls (22,10) and (26,13.5) .. (30,13.5);
            \draw[sceneneutral] (47,16) .. controls (50,14) and (54,7) .. (55,2.8);
            \draw[sceneneutral,mechanism arrow] (55,2.8) .. controls (59,4.4) and (64,4.4) .. (68,3.2);
        \or 
            \node at (31,17) {speed};
            \node at (66,17) {limit};
            \draw[sceneneutral!20,line width=3pt] (8,9) -- (68,9);
            \draw[mechanism guide] (8,6) -- (8,12);
            \draw[mechanism arrow] (8,9) -- (61,9);
            \draw[sceneaccent,line width=0.9pt] (68,5.5) -- (68,12.5);
            \node at (8,1.5) {0};
            \node at (68,1.5) {2.2};
        \or 
            \fill[sceneguide!45] (29,1) rectangle (47,11.5);
            \draw[mechanism guide,densely dashed] (29,1) -- (29,11.5);
            \draw[mechanism guide,densely dashed] (47,1) -- (47,11.5);
            \draw[mechanism arrow] (6,6) -- (24,6);
            \draw[mechanism arrow] (33,6) -- (40,6);
            \draw[mechanism arrow] (52,6) -- (68,6);
            \node at (19,15) {$\times0.4$};
            \node at (58,15) {$\times2.5$};
        \or 
            \fill[sceneneutral!18] (12,1) rectangle (64,11);
            \draw[sceneneutral,line width=1.5pt] (12,1) -- (64,1);
            \fill[sceneneutral] (17,6) circle (2.1);
            \draw[mechanism arrow,sceneneutral] (22,6) -- (35,6);
            \draw[mechanism arrow,sceneneutral] (39,6) -- (61,6);
            \node at (38,15) {$v\!\uparrow$};
        \or 
            \node at (38,15) {separation: $v\to1.8v$};
            \fill[sceneneutral] (34,5) circle (2.5);
            \fill[sceneneutral] (42,5) circle (2.5);
            \draw[mechanism arrow] (29,5) -- (10,5);
            \draw[mechanism arrow] (47,5) -- (66,5);
        \or 
            \begin{scope}[every node/.style={circle,draw=none,text=white,
                inner sep=0pt,minimum size=9.5pt,
                font=\sffamily\bfseries\fontsize{9}{10.5}\selectfont}]
                \node[fill=scenered] at (13,9.25) {A};
                \node[fill=sceneblue] at (22.5,9.25) {B};
                \node[fill=sceneblue] at (52,9.25) {A};
                \node[fill=scenered] at (64,9.25) {B};
            \end{scope}
            \draw[mechanism arrow] (32,9.25) -- (44,9.25);
        \or 
            \node at (9,8) {1};
            \node at (21,8) {2};
            \node[circle,fill=sceneaccent,text=white,inner sep=0pt,minimum size=9.5pt,
                font=\bfseries\fontsize{9}{10.5}\selectfont] at (34,8) {3};
            \draw[mechanism arrow] (42,8) -- (49,8);
            \begin{scope}[every node/.style={circle,fill=sceneneutral,
                text=white,inner sep=0pt,minimum size=9.5pt,
                font=\sffamily\bfseries\fontsize{9}{10.5}\selectfont}]
                \node at (57,5) {A};
                \node at (69,5) {B};
            \end{scope}
            \draw[mechanism arrow] (57,10.5) -- (57,18.5);
            \draw[mechanism arrow] (69,10.5) -- (69,18.5);
        \or 
            \node at (38,15) {12 s};
            \node[circle,fill=sceneneutral,text=white,inner sep=0pt,
                minimum size=9.5pt,
                font=\sffamily\bfseries\fontsize{9}{10.5}\selectfont]
                at (8,5.5) {A};
            \draw[mechanism arrow] (14,5.5) -- (26,5.5);
            \draw[mechanism guide,densely dotted] (28,5.5) -- (47,5.5);
            \draw[fill=scenebackground] (38,5.5) circle (4.1);
            \draw (38,8.2) -- (38,5.5) -- (40.2,4.5);
            \node[circle,fill=sceneneutral,text=white,inner sep=0pt,
                minimum size=9.5pt,
                font=\sffamily\bfseries\fontsize{9}{10.5}\selectfont]
                at (54,5.5) {B};
            \draw[mechanism arrow] (60,5.5) -- (72,5.5);
        \or 
            \node[inner sep=0pt,outer sep=0pt] (pauseTrigger)
                at (23,15) {launch};
            \node[inner sep=0pt,outer sep=0pt] (resumeDelay)
                at (54,15) {10~s};
            \node at (7,5) {on};
            \node at (38,5) {off};
            \node at (69,5) {on};
            \draw[mechanism guide] ([yshift=-0.5pt]pauseTrigger.south) -- (23,5);
            \draw[mechanism guide] ([yshift=-0.5pt]resumeDelay.south) -- (54,5);
            \draw[mechanism arrow] (15,5) -- (31,5);
            \draw[mechanism arrow] (47,5) -- (61,5);
        \fi
        \end{pgfinterruptboundingbox}
    \end{tikzpicture}%
}
\relax\space
    \newcommand{\scenecard}[2]{%
        \begingroup
        \ifnum#1=1\relax
            \def\cardfill{scenecontrol}%
            \def\cardedge{scenecontroledge}%
            \def\cardbadgefill{scenecontrol}%
            \def\cardbadgeink{scenecontroledge}%
        \else
            \def\cardfill{scenebackground}%
            \def\cardedge{sceneborder}%
            \def\cardbadgefill{scenebadge}%
            \def\cardbadgeink{scenebadgeink}%
        \fi
        \pgfmathsetlengthmacro{\scenewidth}{\linewidth-0.35pt}%
        \begin{tikzpicture}[x=1pt,y=1pt,baseline=(sceneheading.base)]
        \shade[top color=\cardfill!15!white,
            middle color=\cardfill!58!white,bottom color=\cardfill,
            opacity=0.85,rounded corners=2.4pt]
            (0,0) rectangle (\scenewidth,47.3);
        \begin{scope}
            \clip[rounded corners=2.4pt] (0,0) rectangle (\scenewidth,47.3);
            \fill[white,opacity=0.12] (0,17) -- (0,34)
                -- (\scenewidth,47.3) -- (\scenewidth,39) -- cycle;
        \end{scope}
        \draw[draw=\cardedge!72!white,rounded corners=2.4pt,
            line width=0.35pt] (0,0) rectangle (\scenewidth,47.3);
        \draw[draw=white,opacity=0.85,line width=0.65pt,rounded corners=1.8pt]
            (0.85,0.85) rectangle (\dimexpr\scenewidth-0.85pt\relax,46.45);
        \node[anchor=west,top color=\cardbadgefill!15!white,
            bottom color=\cardbadgefill,draw=white,line width=0.2pt,
            outer sep=0.2pt,
            rounded corners=1.5pt,
            minimum width=13pt,minimum height=9pt,inner sep=0pt,
            font=\sffamily\bfseries\fontsize{9}{10.5}\selectfont,
            text=\cardbadgeink] at (3.5,41.7) {#1};
        \node[anchor=base,inner sep=0pt,outer sep=0pt,
            font=\bfseries\fontsize{9}{10.5}\selectfont,text=sceneink]
            (sceneheading) at (0.5*\scenewidth,29.3) {#2};
        \draw[draw=sceneguide!70,line width=0.35pt,
            dash pattern=on 1.3pt off 1.3pt]
            (4,25.2) -- (\dimexpr\scenewidth-4pt\relax,25.2);
        \node[inner sep=0pt,outer sep=0pt]
            at (0.5*\scenewidth,12.3) {\scenesymbol{#1}};
        \end{tikzpicture}%
        \endgroup
    }
    \setlength{\tabcolsep}{0pt}
    \renewcommand{\arraystretch}{0.95}
    \begin{tabularx}{1.0\linewidth}{@{}X@{\hspace{6pt}}X@{\hspace{6pt}}X@{\hspace{6pt}}X@{\hspace{6pt}}X@{}}
        \scenecard{01}{Normal control}
        & \scenecard{02}{Red attraction}
        & \scenecard{03}{Mass-dependent fall}
        & \scenecard{04}{Oscillating gravity}
        & \scenecard{05}{High-speed braking} \\
        \addlinespace[6pt]
        \scenecard{06}{Automatic side force}
        & \scenecard{07}{Tile rebounds}
        & \scenecard{08}{Speed has a limit}
        & \scenecard{09}{Slows in, speeds out}
        & \scenecard{10}{Acceleration zone} \\
        \addlinespace[6pt]
        \scenecard{11}{Collision boost}
        & \scenecard{12}{Contact color swap}
        & \scenecard{13}{Third-contact jump}
        & \scenecard{14}{Velocity copy}
        & \scenecard{15}{Northward force} \\
    \end{tabularx}
    \endgroup
    \caption{Scene~01 uses ordinary physics. Scenes~02--15 each have one altered physical law. Laws are in App.~\ref{apx:scenes}.}
    \label{tab:scene-overview}
\end{table*}
\relax\space
\section{Introduction}
\label{sec:introduction}
Large language models (LLMs) aid scientific discovery through hypothesis generation, experiment design, and evidence analysis.
Scientific discovery often begins with an unexpected observation that prompts a new research question.
To pursue such inquiry autonomously, LLMs must decide what to investigate and test their explanations through experiments.
This ability could help scientists investigate unexpected phenomena and pursue questions they might otherwise overlook.
LongChang's name reflects independent inquiry, inspired by the philosopher Wang Yangming's introspection in a cave at Longchang, where he reconsidered how to seek knowledge and reached an insight that helped shape his own School of Mind~\citep{qian1563yangmingnianpu}.

\begin{figure}[t]
    \centering
    \includegraphics[width=0.90\columnwidth]{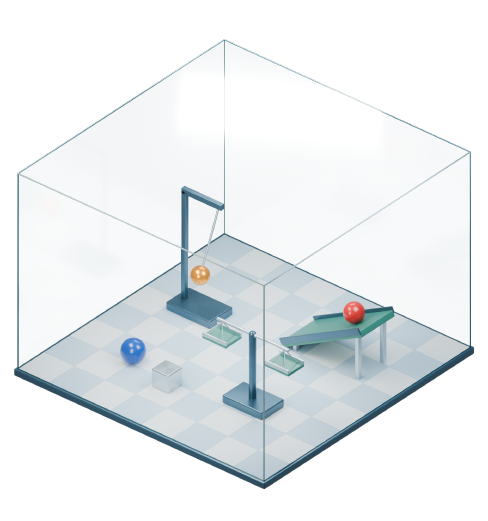}
    \caption{LongChang's glass room. Explorers manipulate objects and build tools. Other LLMs evolve scenes using exploration records without human intervention.}
    \label{fig:discovery-teaser}
\end{figure}

Physical law discovery involves inferring an environment's laws from observations and experiments.
PhysGym~\citep{chen2025physgym}, NewtonBench~\citep{zheng2026newtonbench}, and DiscoverPhysics~\citep{wiemann2026discoverphysics} assign law-discovery tasks.
Tab.~\ref{tab:benchmark-comparison} compares their exploration guidance and experiments with those of LongChang.
Without a research task or hints about altered physical laws, an LLM must notice phenomena worth investigating, choose a research question, and devise experiments to distinguish competing explanations.
Can LLMs turn unexpected observations into research questions, conduct experiments, and discover altered physical laws autonomously?

We introduce LongChang, an automated framework that generates simulated worlds with altered physics, where one or more laws differ from ordinary physics.
This counterfactual setting tests whether LLMs independently notice departures from ordinary physics and investigate them without research tasks or hints about the altered physical laws.
For example, slower falls for heavier objects could lead an explorer to ask how mass affects falling and design controlled comparisons to test this link.
These laws are hidden from explorers but available for evaluating their reported explanations.

LongChang's base set in Tab.~\ref{tab:scene-overview} comprises a normal control and 14 scenes with one altered physical law each.
In the virtual ``glass room'' in Fig.~\ref{fig:discovery-teaser}, explorers manipulate objects and build tools that interact with the scene objects within a shared MuJoCo rigid-body simulation~\citep{todorov2012mujoco}.
Separate LLM agents review tool designs and evaluate reports against the altered physical laws and exploration records, requiring both correctness and experimental support for discovery.
Unresolved questions and corrected errors in these records guide LLM scene designers to generate, refine, and compose scenes for further exploration, continuing the entire discovery cycle without human intervention.

Among 12 evaluated LLMs, GPT-6-Astra achieves a discovery rate of 16.7\% under free exploration, while the other 11 score 0\%.
Exploration records show some LLMs leave observed anomalies uninvestigated, while others experiment but only partially identify the altered physical laws, leaving their conditions or applicability unresolved.

We contribute the LongChang framework for the complete discovery cycle and an empirical study of autonomous physical inquiry.
We use model comparisons, controlled interventions, and exploration records to examine how LLMs form research questions and test explanations using experiments.
\relax\space
\begin{table}[t]
    \centering
    \fontsize{9}{10.5}\selectfont
    \setlength{\tabcolsep}{4.5pt}
    \renewcommand{\arraystretch}{0.95}
    \begin{tabular*}{\linewidth}{@{\extracolsep{\fill}}lcccc@{}}
        \toprule
        \multicolumn{1}{@{}l!{\textcolor{black!35}{\rule[-6.5pt]{0.3pt}{18pt}}}}{\textbf{Closest work}}
        & \multicolumn{1}{c!{\textcolor{black!35}{\rule[-6.5pt]{0.3pt}{18pt}}}}{\begin{tabular}[c]{@{}c@{}}\textbf{No}\\\textbf{task}\end{tabular}}
        & \multicolumn{1}{c!{\textcolor{black!35}{\rule[-6.5pt]{0.3pt}{18pt}}}}{\begin{tabular}[c]{@{}c@{}}\textbf{Interactive}\\\textbf{3D scenes}\end{tabular}}
        & \multicolumn{1}{c!{\textcolor{black!35}{\rule[-6.5pt]{0.3pt}{18pt}}}}{\begin{tabular}[c]{@{}c@{}}\textbf{Tool}\\\textbf{design}\end{tabular}}
        & \begin{tabular}[c]{@{}c@{}}\textbf{Scene}\\\textbf{evolution}\end{tabular} \\
        \midrule
        PhysGym & \ding{55} & \ding{55} & \ding{55} & \ding{55} \\
        \addlinespace[2pt]
        NewtonBench & \ding{55} & \ding{55} & \ding{55} & \ding{55} \\
        \addlinespace[2pt]
        DiscoverPhysics & \ding{55} & \ding{55} & \ding{55} & \ding{55} \\
        \midrule
        \textbf{LongChang} & \ding{51} & \ding{51} & \ding{51} & \ding{51} \\
        \bottomrule
    \end{tabular*}
    \caption{LongChang compared with the closest work: PhysGym~\citep{chen2025physgym}, NewtonBench~\citep{zheng2026newtonbench}, and DiscoverPhysics~\citep{wiemann2026discoverphysics}. No task: no assigned research goal. Interactive 3D scenes: physical worlds in MuJoCo~\citep{todorov2012mujoco}. Tool design: devices built to interact with scene objects. Scene evolution: automatically generating and revising scenes guided by exploration records.}
    \label{tab:benchmark-comparison}
\end{table}

\section{Related Work}
\label{sec:related-work}

\subsection{Autonomous Scientific Discovery}
\label{sec:related-discovery}
\label{sec:related-task-free}

PhysGym~\citep{chen2025physgym} and NewtonBench~\citep{zheng2026newtonbench} ask LLMs to recover hidden equations through numerical queries, while DiscoverPhysics~\citep{wiemann2026discoverphysics} studies law discovery in particle simulations.
These benchmarks specify discovery goals before experimentation, as summarized in Tab.~\ref{tab:benchmark-comparison}.
Other protocols require explicit initial hypotheses and a minimum number of experiments~\citep{naegele2026sciexplorer} or supply the anomaly to explain~\citep{geng2026eurekabench}.
Open-ended systems let LLMs select questions within an assigned research role, using supplied datasets~\citep{agarwal2025autodiscovery,chen2026statefuldiscovery,yao2025scilink} or an optical research setting~\citep{zeng2026socratic,yang2026qiushi}.
LongChang tests whether LLMs notice unusual effects of altered physical laws, initiate investigations, and establish these laws experimentally, without a preassigned research task or hints about these laws.

Left Alone~\citep{szeider2025leftalone} studies self-directed behavior without an assigned task or a simulated physical world.
Explore-then-Act in Look Before You Leap~\citep{ye2026look} studies task-free exploration, including in ScienceWorld, with modeling instructions and a warning that expectations may fail.
It evaluates exploration and downstream task performance, rather than experimentally supported discovery of altered physical laws by the explorer.

Both WorldTest~\citep{warrier2026worldtest} and OneLife~\citep{khan2026onelife} examine predictive world models and explicitly ask explorers to learn environment rules.
LongChang instead asks whether unexpected physical observations prompt LLMs to pose their own questions and establish altered physical laws through experiments, without research tasks or advance hints.
Tab.~\ref{tab:benchmark-comparison} therefore focuses on frameworks for physical law discovery.

\begin{figure*}[t]
    \centering
    \includegraphics[trim=36bp 0 36bp 0,clip,width=1.0\textwidth]{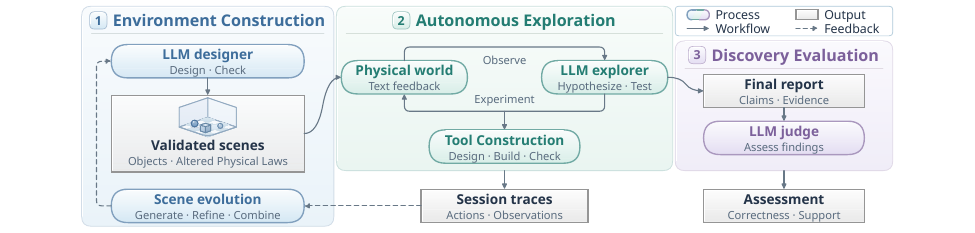}
    \caption{LongChang's automated cycle of scene construction, free exploration, assessment, and scene evolution.}
    \label{fig:task-overview}
\end{figure*}

\subsection{Tool Use and Design}
\label{sec:related-exploration}

DiscoveryWorld supplies instruments and abstracted object interactions in a grid world~\citep{jansen2024discoveryworld}, while ChemWorld exposes chemical procedures through programmable modules~\citep{qiu2026chemworld}.
Tool-design systems optimize tool geometry and use for a specified manipulation task~\citep{lin2025robotsmith} or propose detector layouts under given performance goals and resource constraints~\citep{zoccheddu2026instrument}.
LongChang ties tool construction to the explorer's own investigation: LLMs decide whether a tool is needed, what to build, and how to use it to test a hypothesis.
The explorer's constructed tools interact physically with the other objects in the same simulated world.

\subsection{Environment Generation and Evolution}
\label{sec:related-evolution}

Environment-generation methods create tasks and rewards for agent training and adapt environments based on performance or trajectory feedback~\citep{faldor2024omniepic,kang2026simworldstudio}.
Programmable chemical worlds support controlled changes to private laws while preserving public experimental conditions~\citep{qiu2026chemworld}.
Trajectory-based benchmark construction uses tool-use records to create new tasks with solutions checked by re-execution~\citep{guo2025trace}.
In LongChang, LLM scene designers use unresolved questions and corrected errors in exploration records to generate, refine, and compose scenes.
Explorer LLMs choose observations to investigate, while separate LLM agents review tools and assess reports, continuing this discovery process without human intervention.

\section{LongChang}
\label{sec:task}

Fig.~\ref{fig:task-overview} presents LongChang's automated cycle of physical law discovery: environment construction (Sec.~\ref{sec:environment-construction}), autonomous exploration (Sec.~\ref{sec:behavior}), and discovery evaluation (Sec.~\ref{sec:evaluation}).
The evaluated LLMs serve as explorers, while separate LLM agents design and review scenes, review apparatus, and evaluate reports.
LLMs carry out the entire loop without human intervention, using exploration records to guide the subsequent evolution of scenes.

\subsection{Environment Construction}
\label{sec:environment-construction}

\begin{figure*}[t]
    \centering
    \begin{minipage}[t]{\columnwidth}
        \centering
        \includegraphics[scale=1,keepaspectratio,width=1.0\linewidth]{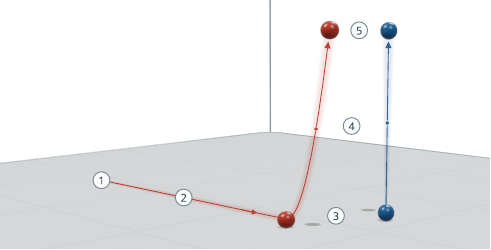}
        \caption{An example of an altered physical law. When the two balls are close, an upward force makes them rise together. Labels 1--5 identify successive snapshots.}
        \label{fig:scene-examples}
    \end{minipage}\hfill
    \begin{minipage}[t]{\columnwidth}
        \centering
        \includegraphics[scale=1,keepaspectratio,width=1.0\linewidth]{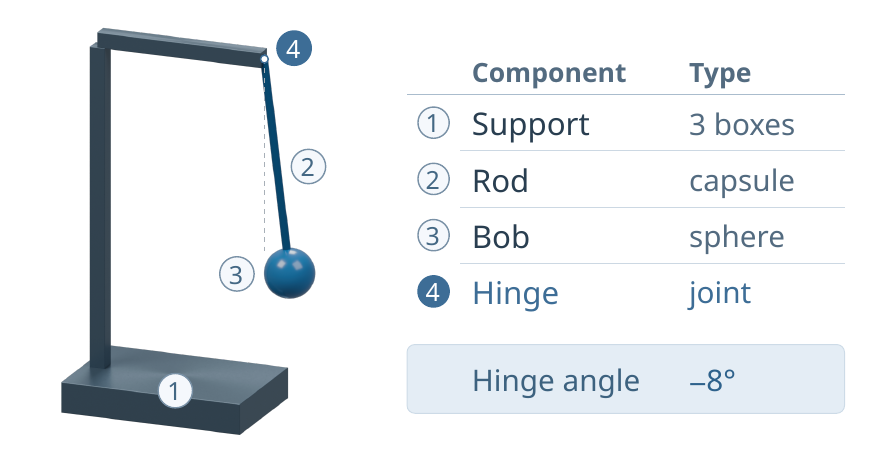}
        \caption{An example of an LLM-built tool. The pendulum uses its hinge angle to test whether local gravity is tilted. The instrument reads $-8^\circ$ in the illustrated state.}
        \label{fig:apparatus-example}
    \end{minipage}
\end{figure*}

\noindent\textbf{Physical Environment. }
An LLM designer specifies each scene's objects and geometry, physical program, and any altered physical laws.
The framework compiles these definitions into MuJoCo simulations~\citep{todorov2012mujoco} containing manipulable balls and boxes, with colored floor tiles in some scenes.
Each base scene with altered physics modifies one physical law, producing effects that depend on object properties, spatial relations, contact events, or past actions.
Fig.~\ref{fig:scene-examples} illustrates an altered physical law: when the balls are close, both rise under an upward force without an upward launch.

The designer records each altered physical law's target objects, conditions, and observable effects for evaluation, withholding this information from the explorer.
LongChang's base set in Tab.~\ref{tab:scene-overview} comprises a normal control and 14 scenes with one altered physical law each.
The normal control assesses whether LLMs ground their conclusions in evidence when no physical law has been altered.
App.~\ref{apx:scenes} details ordinary physics and each altered physical law's scope, conditions, and parameters.

\noindent\textbf{Scene Evolution. }
To extend this set, an LLM designer proposes scene changes aimed at more challenging discovery tasks.
The designer receives source scenes and any available summaries of earlier explorations.
These summaries identify tested conditions, unresolved questions, and explanations supported or corrected by later observations.
The designer uses three operations: generation creates new laws and scenes from source mechanisms, refinement modifies an existing scene, and composition combines source laws in a shared environment.
Changes guided by these summaries must cite supporting exploration evidence.
Automated tests check simulated behavior against the altered physical laws, including interactions with constructed apparatus.
An LLM reviewer checks the program and test results against these laws.
Failed compilation, tests, or review trigger automatic revision requests to the designer.
Only accepted scenes are used in subsequent free-exploration assessments.

\subsection{Autonomous Exploration}
\label{sec:behavior}

The explorer chooses which observations to investigate, formulates its own research questions, and tests explanations through physical interventions.
The initial prompt asks for free exploration without a research task or hints about the altered physics.
Structured observations give object positions, motion, and instrument readings, with distances in bricks and elapsed time in counts.
The explorer can push or throw objects, grasp and move them, release them, wait, and mark reference locations.
App.~\ref{apx:action-parameters} documents the complete action interface.

When an investigation needs additional control or measurement, the explorer can design apparatus such as the pendulum in Fig.~\ref{fig:apparatus-example}.
It specifies the device's geometry and mechanical connections, including joints and springs.
An LLM reviewer checks the compiled design for physical validity and returns feedback on rejections.
The explorer can revise a rejected design and submit it again, with successful designs installed into the current scene.
Once installed, the explorer can place objects on the apparatus or adjust its joints and receive updated motion readings with the next observation.

At each turn, the explorer records its observations and action rationale, then acts and receives updated observations.
The cycle continues until it stops or reaches the turn limit, retaining notes, actions, observations, and feedback from all attempts.

The final report summarizes the observations and explanations from this record.
Each conclusion must cite supporting turns and use only interpretations already recorded in the notes, preserving the uncertainty stated there.
The explorer may also report that it found no anomaly during exploration.

\subsection{Discovery Evaluation}
\label{sec:evaluation}

An LLM judge evaluates reports against altered physical laws and full exploration records, requiring law correctness (C) and evidence support (S) for discovery.
Law correctness requires identifying each altered physical law, its target objects, conditions, triggers, and constants identifiable in the reported units.
Proposed mechanisms may differ if they yield the same observable predictions under allowed interventions.
Evidence support requires recorded experiments to justify all claims and reasoning at the stated level of certainty.
Correct guesses may pass only C. Supported ordinary observations or partial findings may meet S alone.

\subsection{Experimental Setup}
\label{sec:exp-setup}

\noindent\textbf{Main evaluation. }
All LLMs explore the same 15 scenes from fixed initial states for up to 100 turns.

\noindent\textbf{LLMs. }
We evaluate the 12 LLMs in Tab.~\ref{tab:main-results}.
GPT-6.1-Sol judges the correctness and experimental support of their reports.
GPT-6-Astra reviews apparatus and generates and reviews new scenes.
App.~\ref{apx:prompts} lists the prompts used across the pipeline.

\noindent\textbf{Metrics. }
For scenes with altered physics, C and S show the percentages of runs passing the relevant check.
Discovery rate (D) is the percentage passing both checks.
Rejection count (R) is the number of rejected normal-control reports, including those with false claims of anomalies or unsupported laws.

\noindent\textbf{Exploration statistics. }
Across all scenes, exploration length (L) is the mean number of turns used, and early stopping (ES) is the percentage of runs ending before the turn limit.
Action failure (AF) is the percentage of turns with failed actions.
Construction success (CS) counts explorations with at least one successful build, and construction failure (CF) gives the percentage of failed build attempts.
These statistics distinguish budget use, access to working apparatus, and action execution reliability.

\begin{table*}[t]
\centering
\fontsize{9}{10.5}\selectfont
\setlength{\tabcolsep}{5pt}
\renewcommand{\arraystretch}{0.95}
\begin{tabular*}{1.0\linewidth}{@{\extracolsep{\fill}}lrrrrrrrrr@{}}
\toprule
& \multicolumn{3}{c}{Discovery (\%)} & Control & \multicolumn{5}{c}{Exploration} \\
\cmidrule(lr){2-4}\cmidrule(lr){5-5}\cmidrule(l){6-10}
Model & D & C & S & R & L & ES (\%) & CS & CF (\%) & AF (\%) \\
\midrule
GPT-5.6-Luna & 0.0 & 0.0 & 38.1 & 0 & 36.7 & 95.6 & 0 & -- & 0.8 \\
GPT-6-Astra & 16.7 & 19.0 & 97.6 & 0 & 39.3 & 100.0 & 11 & 48.0 & 2.1 \\
Claude-Opus-5 & 0.0 & 16.7 & 0.0 & 0 & 85.8 & 100.0 & 44 & 66.6 & 18.2 \\
Gemini-3.7-Flash & 0.0 & 0.0 & 19.0 & 0 & 10.9 & 100.0 & 1 & 75.0 & 6.1 \\
GLM-5.3 & 0.0 & 4.8 & 0.0 & 3 & 91.6 & 97.8 & 45 & 64.0 & 16.8 \\
GLM-5.3-Flash & 0.0 & 4.8 & 0.0 & 1 & 87.6 & 95.6 & 45 & 71.4 & 20.2 \\
Kimi-K3 & 0.0 & 2.4 & 0.0 & 3 & 83.9 & 93.3 & 42 & 82.1 & 32.7 \\
MiMo-V2.6-Flash & 0.0 & 4.8 & 0.0 & 1 & 36.2 & 100.0 & 33 & 71.1 & 24.2 \\
MiMo-V2.6-Pro & 0.0 & 7.1 & 0.0 & 2 & 42.3 & 100.0 & 35 & 70.1 & 19.5 \\
DeepSeek-Flash & 0.0 & 0.0 & 0.0 & 2 & 45.6 & 100.0 & 42 & 77.1 & 19.2 \\
Qwen3-Max & 0.0 & 0.0 & 9.5 & 0 & 26.7 & 100.0 & 16 & 63.8 & 6.4 \\
Grok-4.7 & 0.0 & 0.0 & 2.4 & 1 & 85.2 & 24.4 & 2 & 88.5 & 12.2 \\
\bottomrule
\end{tabular*}
\caption{Discovery performance and exploration behavior of 12 LLMs in LongChang, with metrics from Sec.~\ref{sec:exp-setup}.}
\label{tab:main-results}
\end{table*}
\relax\space

\begin{figure*}[!t]
    \centering
    \includegraphics[width=1.0\textwidth]{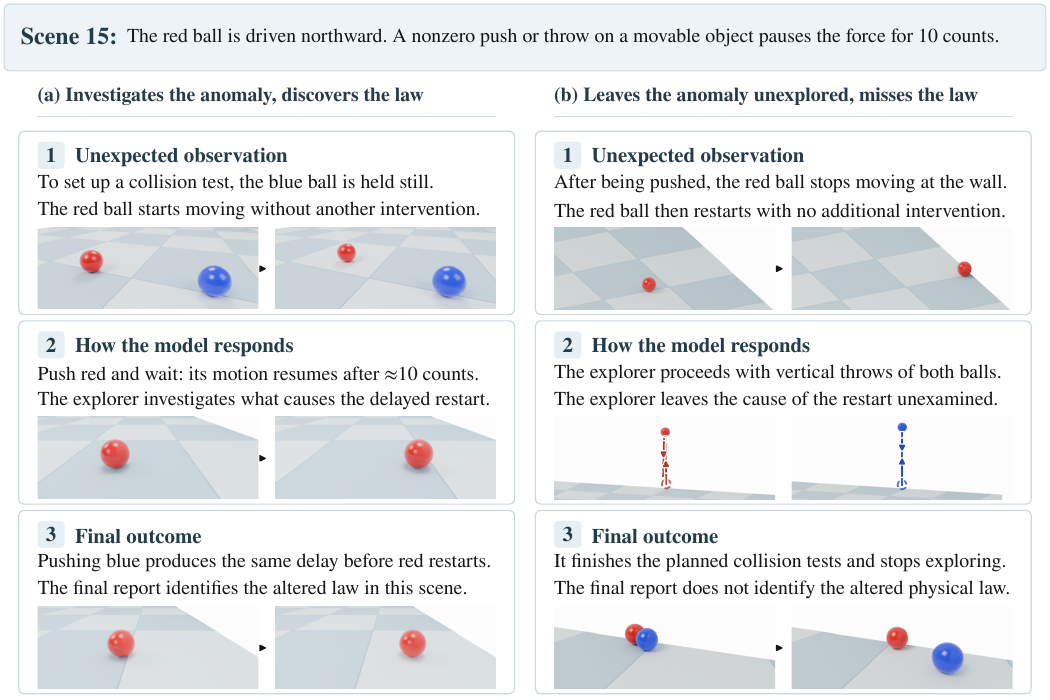}
    \caption{Case study of research question formation in Scene~15. (a) GPT-6-Astra investigates red's delayed restart and discovers the altered physical law. (b) Gemini-3.7-Flash observes the restart but still follows its original plan.}
    \label{fig:research-question}
\end{figure*}

\begin{figure}[!t]
    \centering
    \includegraphics[width=1.0\columnwidth]{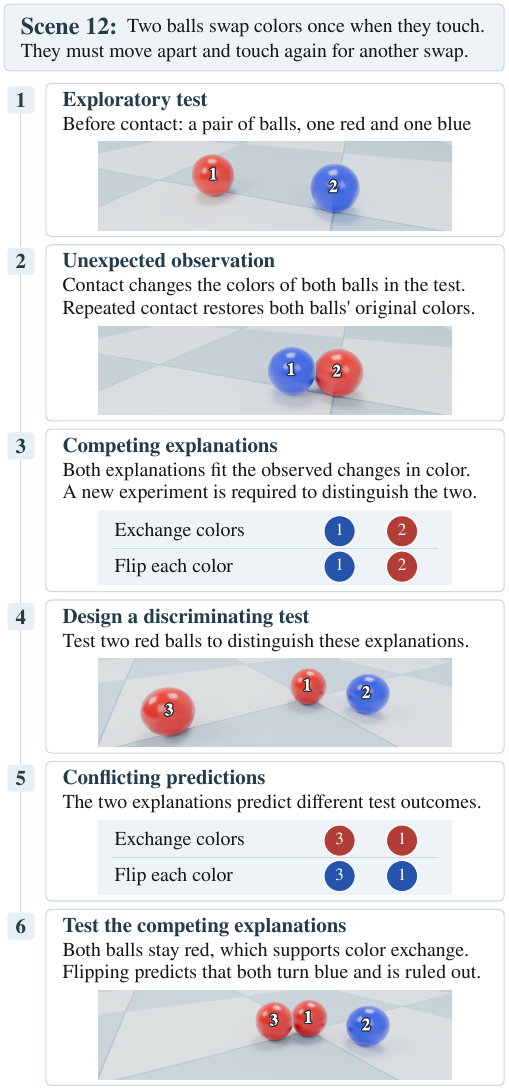}
    \caption{Case study of testing competing explanations.}
    \label{fig:case-studies}
\end{figure}
\relax\space
\begin{figure}[!t]
    \centering
    \includegraphics[width=1.0\columnwidth]{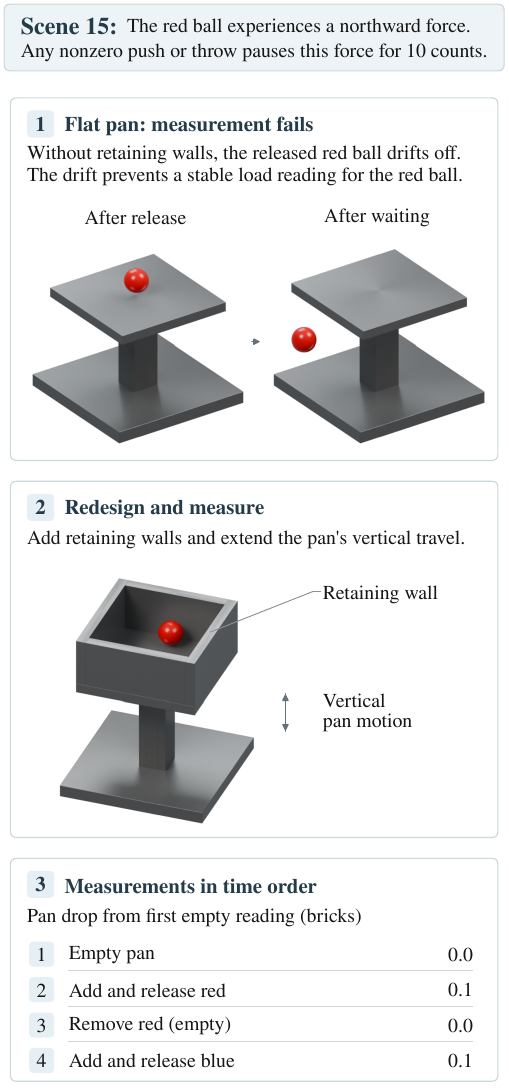}
    \caption{Case study of tool construction and redesign.}
    \label{fig:tool-case}
\end{figure}
\relax\space

\section{Results and Analysis}
\label{sec:analysis}
\label{sec:results}

Tab.~\ref{tab:main-results} shows that GPT-6-Astra achieves a discovery rate of 16.7\% under free exploration, while the other 11 LLMs score 0\%.
Its evidence support rate reaches 97.6\%, but law correctness is only 19.0\%: supported observations and partial findings often leave the altered physical laws unresolved.
For example, a Scene~02 report supports attraction between the tested red balls without establishing its distance dependence or applicability to other red spheres.
Claude-Opus-5 shows the opposite pattern, identifying the altered physical laws in 16.7\% of runs while no report passes the evidence support check.
These results separate correct identification of a law from experimentally supported discovery.

Exploration statistics reveal obstacles during investigation.
Gemini-3.7-Flash stops early in every run, averaging only 10.9 turns and leaving little opportunity for follow-up experiments.
The GLM models use about 90 turns and build working apparatus in every run, yet make no discoveries.
They also fail in more than 60\% of build attempts, with failed actions accounting for 16.8--20.2\% of turns.
Successful construction can thus coexist with repeated unsuccessful attempts.
Grok-4.7 stops early least often but makes no discoveries, so persistence and working devices alone do not ensure discovery.

\subsection{Forming Research Questions}
\label{sec:analysis-questions}

We study whether explorers turn unexpected observations into research questions and pursue them experimentally.
Fig.~\ref{fig:research-question} compares two Scene~15 explorations with the same initial state and prompt.
When the red ball starts moving while the blue ball is held, GPT-6-Astra postpones its planned collision test to investigate.
It measures the restart delay and tests whether pushing blue also resets it, discovering that pushes or throws pause red's northward force for ten counts.
Gemini-3.7-Flash observes a similar restart but continues its planned tests, leaving the unexpected motion unexamined.

We test guidance by giving GPT-6-Astra a law-discovery task, noting that physical laws may be altered, and supplying a scene-specific clue, with other settings fixed.
In Tab.~\ref{tab:controlled-comparisons}, discovery rises from 16.7\% to 28.6\%, while evidence support remains unchanged.
Guidance improves law identification when experimental support is already high.
Mean exploration length falls from 39.3 to 30.9 turns, yielding more discoveries with fewer interactions.

\begin{table}[!t]
\centering
\fontsize{9}{10.5}\selectfont
\setlength{\tabcolsep}{3pt}
\renewcommand{\arraystretch}{0.95}
\begin{tabular*}{1.0\linewidth}{@{\extracolsep{\fill}}lrrrrr@{}}
\toprule
Condition & D (\%) & C (\%) & S (\%) & L & AF (\%) \\
\midrule
\multicolumn{6}{@{}l}{\textit{Task guidance, 100 turns}} \\
Free exploration & 16.7 & 19.0 & 97.6 & 39.3 & 2.1 \\
Task-guided & 28.6 & 31.0 & 97.6 & 30.9 & 2.7 \\
\midrule
\multicolumn{6}{@{}l}{\textit{Turn budget}} \\
100 turns & 16.7 & 16.7 & 100.0 & 42.2 & 2.7 \\
500 turns & 14.3 & 14.3 & 100.0 & 41.4 & 3.6 \\
\midrule
\multicolumn{6}{@{}l}{\textit{Apparatus access, 100 turns}} \\
Enabled & 16.7 & 19.0 & 97.6 & 39.3 & 2.1 \\
Disabled & 4.8 & 4.8 & 100.0 & 38.7 & 0.9 \\
\bottomrule
\end{tabular*}
\caption{Effects of task guidance, turn budget, and apparatus access on discovery and exploration behavior.}
\label{tab:controlled-comparisons}
\end{table}
\relax\space

\subsection{Validating Scientific Claims}
\label{sec:analysis-validation}

Distinguishing competing explanations requires experiments for which they predict different outcomes.
Fig.~\ref{fig:case-studies} illustrates this process in Scene~12.
Two successive red--blue collisions change and restore the balls' colors, consistent with both color exchange and independent red--blue flipping.
GPT-6-Astra tests two red balls, for which exchange predicts that both remain red and flipping predicts that both turn blue.
Both balls remain red, supporting exchange and ruling out independent flipping.

The normal control tests whether LLMs overinterpret observations under ordinary physics.
Across LLMs, 36.1\% of reports are rejected for false claims of altered physical laws or unsupported explanations of ordinary physics.
For example, DeepSeek-Flash attributes a ball's longer travel on the decorative gray tile to lower friction after comparing equal pushes.
The ball is already moving before each push, so different starting speeds could explain the difference in distance.
A fair comparison would start the ball from rest before each push.

\subsection{Using Experimental Resources}
\label{sec:analysis-resources}

We examine whether larger budgets or apparatus access helps LLMs resolve unanswered questions.
We compare budgets of 100 and 500 turns under free exploration without context compaction, with other settings matched.
Increasing the limit to 500 turns does not improve discovery: GPT-6-Astra uses about 40 turns under either limit and stops early, sometimes leaving questions unresolved.
Extra turns provide little value if explorers stop early.

The apparatus comparison uses a 100-turn limit.
Disabling construction removes its actions and instructions but retains basic object manipulation.
With construction enabled, discovery rises from 4.8\% to 16.7\% and law correctness from 4.8\% to 19.0\% in Tab.~\ref{tab:controlled-comparisons}, although action failures also increase.
The explorer chooses whether to build a tool, so these results concern access to construction rather than the contribution of tool use in each run.

Fig.~\ref{fig:tool-case} illustrates the construction and redesign of a weighing platform in Scene~15.
GPT-6-Astra builds a flat pan to compare the balls' loads, but the red ball drifts off after release.
It adds retaining walls and increases the pan's vertical travel, obtaining stable readings after releasing each ball.

\section{Scene Evolution}
\label{sec:scene-evolution}
\label{sec:exp-evolution}

Exploration records can also inform what LLMs investigate in subsequent scenes.
We study whether corrected errors from successful explorations can guide generation, refinement, and composition of new discovery challenges.
Each operation modifies a mechanism involved in the same error to create a scene with altered physics for further exploration.

We review successful GPT-6-Astra explorations for explanations corrected after later observations.
In Scene~15, red stops after a push, prompting the inference that no persistent northward force exists.
When red restarts, the explorer revises this explanation to account for a temporary pause in the force.
The designer receives this error, its correction, and the source laws, and uses the same evidence for all three operations.
Tab.~\ref{tab:scene-evolution-cases} shows how the three designs retain the pause while modifying these laws.

\noindent\textbf{Generation. }
The designer replaces the fixed northward force on red with mutual repulsion between red and blue, making the force depend on their relative positions.
Swapping their positions at matched times after a launch should reverse the forces, distinguishing repulsion from a fixed directional force.

\noindent\textbf{Refinement. }
The designer introduces a spatial condition blocking red's northward force while blue is inside the gray square, so waiting longer than ten seconds may still not restore it.
After the pause expires, the planned test moves blue by hand into or out of the square without resetting the timer.

\noindent\textbf{Composition. }
The designer combines the paused northward force with Scene~14's delayed motion copying: green takes red's launch velocity twelve seconds later.
The planned comparison launches blue after red, postponing red's force recovery while leaving green's scheduled velocity copy unchanged.
The explorer must identify both altered physical laws and their distinct triggers and timing.

\begin{table*}[!t]
\centering
\fontsize{9}{10.5}\selectfont
\setlength{\tabcolsep}{7pt}
\renewcommand{\arraystretch}{1.0}
\renewcommand{\tabularxcolumn}[1]{p{#1}}
\definecolor{evoSceneFill}{HTML}{EDF3F6}
\definecolor{evoSceneEdge}{HTML}{CBD8DE}
\definecolor{evoCardEdge}{HTML}{CBD8DE}
\definecolor{evoExampleFill}{HTML}{EDF3F6}
\definecolor{evoHeading}{HTML}{243C49}
\begin{tikzpicture}[font=\fontsize{9}{10.5}\selectfont]
\path[use as bounding box] (0pt,0pt) rectangle (\linewidth,37pt);
\draw[rounded corners=2.7pt,fill=evoSceneFill,draw=evoSceneEdge,line width=0.5pt] (0pt,20pt) rectangle (\linewidth,37pt);
\node[anchor=west,inner sep=0pt] at (6pt,28.5pt) {\textcolor{evoHeading}{\textbf{Scene 15:}} A nonzero push or throw pauses the red ball's northward force for 10~s.};
\draw[rounded corners=2.7pt,fill=white,draw=evoCardEdge,line width=0.5pt] (0pt,0pt) rectangle (\linewidth,17pt);
\node[anchor=west,inner sep=0pt] at (6pt,8.5pt) {\textcolor{evoHeading}{\textbf{Corrected error.}} The explorer initially interpreted this pause as evidence of \uline{no lasting force}.};
\end{tikzpicture}
\par
\begin{tabularx}{\linewidth}{@{}>{\raggedright\arraybackslash}p{0.44\linewidth}>{\raggedright\arraybackslash}X@{}}
\toprule
\makebox[\linewidth][c]{\textcolor{evoHeading}{\textbf{How the scene evolves}}} & \makebox[\linewidth][c]{\textcolor{evoHeading}{\textbf{Altered physical laws}}} \\
\midrule
\multicolumn{2}{@{}l@{}}{\textcolor{evoHeading}{\textbf{Generation}}} \\
\begin{minipage}[t]{\linewidth}\raggedright Replace the northward force with mutual repulsion between the red and blue balls, retaining the 10~s pause.\end{minipage} & \begin{minipage}[t]{\linewidth}
\begin{tikzpicture}[font=\fontsize{9}{10.5}\selectfont,baseline={([yshift=-8pt]current bounding box.north)}]
\path[use as bounding box] (0pt,0pt) rectangle (\linewidth,23pt);
\draw[rounded corners=2.7pt,fill=evoExampleFill,draw=evoCardEdge,line width=0.5pt] (0pt,0pt) rectangle (\linewidth,23pt);
\node[anchor=base,inner sep=0pt,text=black] at (0.5\linewidth,15pt) {The red and blue balls are subject to mutual horizontal repulsion.};
\node[anchor=base,inner sep=0pt,text=black] at (0.5\linewidth,4.5pt) {Any nonzero push or throw pauses this mutual repulsion for 10~s.};
\end{tikzpicture}
\end{minipage} \\
\noalign{\vskip2pt{\color{evoCardEdge}\hrule height0.35pt}\vskip3pt}
\multicolumn{2}{@{}l@{}}{\textcolor{evoHeading}{\textbf{Refinement}}} \\
\begin{minipage}[t]{\linewidth}\raggedright Block the red ball's northward force while the blue ball is inside the gray square, even after the 10~s pause ends.\end{minipage} & \begin{minipage}[t]{\linewidth}
\begin{tikzpicture}[font=\fontsize{9}{10.5}\selectfont,baseline={([yshift=-8pt]current bounding box.north)}]
\path[use as bounding box] (0pt,0pt) rectangle (\linewidth,23pt);
\draw[rounded corners=2.7pt,fill=evoExampleFill,draw=evoCardEdge,line width=0.5pt] (0pt,0pt) rectangle (\linewidth,23pt);
\node[anchor=base,inner sep=0pt,text=black] at (0.5\linewidth,15pt) {Blue inside the gray square blocks the red ball's northward force.};
\node[anchor=base,inner sep=0pt,text=black] at (0.5\linewidth,4.5pt) {Nonzero pushes or throws also pause the northward force for 10~s.};
\end{tikzpicture}
\end{minipage} \\
\noalign{\vskip2pt{\color{evoCardEdge}\hrule height0.35pt}\vskip3pt}
\multicolumn{2}{@{}l@{}}{\textcolor{evoHeading}{\textbf{Composition}}} \\
\begin{minipage}[t]{\linewidth}\raggedright Add Scene 14's rule: the green ball copies red's launch velocity 12~s after a nonzero push or throw acts on red.\end{minipage} & \begin{minipage}[t]{\linewidth}
\begin{tikzpicture}[font=\fontsize{9}{10.5}\selectfont,baseline={([yshift=-8pt]current bounding box.north)}]
\path[use as bounding box] (0pt,0pt) rectangle (\linewidth,23pt);
\draw[rounded corners=2.7pt,fill=evoExampleFill,draw=evoCardEdge,line width=0.5pt] (0pt,0pt) rectangle (\linewidth,23pt);
\node[anchor=base,inner sep=0pt,text=black] at (0.5\linewidth,15pt) {A nonzero push or throw on any object pauses red's force for 10~s.};
\node[anchor=base,inner sep=0pt,text=black] at (0.5\linewidth,4.5pt) {Each red launch makes green copy red's launch velocity 12~s later.};
\end{tikzpicture}
\end{minipage} \\
\bottomrule
\end{tabularx}
\caption{Scene evolution guided by a corrected error in Scene~15, with altered physical laws in three new scenes.}
\label{tab:scene-evolution-cases}
\end{table*}
\relax\space
\begin{table}[t]
\centering
\fontsize{9}{10.5}\selectfont
\setlength{\tabcolsep}{4pt}
\renewcommand{\arraystretch}{0.95}
\begin{tabular*}{\linewidth}{@{\extracolsep{\fill}}lrrr@{}}
\toprule Scene & C (\%) & S (\%) & D (\%) \\\midrule
Scene 15 & 33.3 & 100.0 & 33.3 \\
Scene 14 & 66.7 & 100.0 & 66.7 \\
Generation & 0.0 & 100.0 & 0.0 \\
Refinement & 0.0 & 100.0 & 0.0 \\
Composition & 0.0 & 100.0 & 0.0 \\
\bottomrule
\end{tabular*}
\caption{Discovery rates in source and evolved scenes.}
\label{tab:evolution-evaluation}
\end{table}
\relax\space

We evaluate GPT-6-Astra on source and evolved scenes using the free-exploration settings in Tab.~\ref{tab:main-results}.

Tab.~\ref{tab:evolution-evaluation} reports discovery rates of 33.3\% in Scene~15 and 66.7\% in Scene~14, compared with 0\% in all three evolved scenes.
Evidence support remains at 100\%, but no evolved-scene report passes law correctness.
In the generated scene, an exploration identifies position-dependent repulsion without recovering its pause and delay.
In the refined scene, the explorer observes northward motion without linking it to blue's position.
In the composed scene, it records red's drift and green's intermittent motion without identifying the two timing rules.
These cases use a corrected error to guide scene evolution, revealing unresolved conditions and dependencies in otherwise supported findings.

\section{Conclusion}

LongChang examines whether LLMs can turn unexpected observations into research questions and discover altered physical laws through experiments.
The evaluated LLMs manipulate objects and build tools without research tasks or hints about these laws.
Only one of the twelve evaluated LLMs makes discoveries, revealing the difficulty of turning observations into complete, experimentally supported laws.
Unanswered questions and corrected errors in exploration records guide separate LLM agents to evolve scenes, continuing the discovery cycle without human intervention.
This framework supports research into LLMs that decide what to investigate and design experiments to resolve uncertainties about the physical laws of these worlds.

\section{Limitations}

This study examines whether LLMs initiate investigations and discover altered physical laws in simulated worlds.
Our results show that this task remains challenging for current LLMs.
The altered physical laws provide a clear basis for evaluating discoveries, while many scientific questions in the real world remain unresolved.
Extending the framework to these open questions requires further research on evaluating exploration and its findings in settings where the answers are not yet established.

\section{Ethical Considerations}

LongChang uses no human annotation or testing and evaluates LLMs without human intervention.
The evaluation therefore raises no ethical concerns related to participant privacy or informed consent.


\bibliography{custom}

\appendix
\definecolor{appendixSceneFill}{HTML}{F5F6F7}
\definecolor{appendixSceneEdge}{HTML}{D8DDE0}
\definecolor{appendixCardEdge}{HTML}{D0D5D9}
\lstdefinestyle{appendixprompt}{
    style=paperprompt,
    breakatwhitespace=true,
    framerule=0.45pt,
    rulecolor=\color{appendixCardEdge},
    backgroundcolor=\color{white},
    literate={—}{{\textemdash}}1
}
\relax\space

\section{Appendix}
\label{sec:appendix}

\subsection{Scene Laws}
\label{apx:scenes}

Scene~01 provides the normal control: three movable balls and a decorative gray tile in a room with ordinary rigid-body physics.
Motion follows $\mathbf F=m\mathbf a$ and $\mathbf p=m\mathbf v$, with constant downward gravitational acceleration independent of mass.
Contact with other objects, the floor, and walls produces collisions and friction.
Object colors remain fixed, and the gray tile has no physical effect on motion.

Scenes~02--15 each change one physical law of this setting.
Tab.~\ref{tab:base-scenes} and Tab.~\ref{tab:evolved-scenes} detail the laws and scope of the 15 base and the three evolved scenes.

\subsection{Action Interface}
\label{apx:action-parameters}

Tab.~\ref{tab:action-parameters} specifies the target objects, parameters, and effects of the actions available during exploration.

\subsection{Prompts}
\label{apx:prompts}

The prompts below guide exploration, generate final reports, and evaluate discoveries.
Task guidance supplements the original exploration prompt.

\subsection{AI Assistance}
\label{apx:ai-assistance}

We used AI assistants to improve the wording and clarity of the manuscript and to assist with code implementation and debugging.
We also used GPT-6.1-Sol as an automated judge to assess the correctness and experimental support of exploration reports, as described in Sec.~\ref{sec:evaluation}.
The authors are responsible for the manuscript, code, and findings.

\subsection{Potential Risks}
\label{apx:potential-risks}

LLM judges may endorse unsupported findings and mislead subsequent research.
Our experiments are simulated, but physical use of generated experiments or apparatus could cause harm without independent validation and expert safety review.
Repeated LLM inference and simulation also consume energy, with associated environmental costs.

\begingroup
\setlength{\fboxsep}{5pt}%
\begin{lstlisting}[style=appendixprompt,title={%
    \colorbox{appendixSceneFill}{%
        \parbox{\dimexpr\linewidth-10pt\relax}{%
            \color{black}\normalfont\bfseries\fontsize{8.5}{10}\selectfont Free exploration}}%
}]
Explore freely and report what you learn.

## Basic operations
- PUSH <object> <heading> <impulse>.
- THROW <object> <heading> <elevation> <impulse>.
- WAIT <counts>: advance the world by 1-30 whole counts.
- GRASP <object>: hold.
- MOVE <object> <x> <y> <z>: move held object.
- RELEASE <object>: let go.
- MARK <id> <x> <y>: place a floor reference.

## Parameters and observations
- Heading: degrees (0=east, 90=north) or east/north/west/south; elevation: 0-90 degrees.
- Impulse: 0-10, added to current motion. Positions: bricks.
- RELEASE a held target before PUSH/THROW.
- Object motion is in events[name].trajectory: count, position, state. Counts are relative to the action start and may be fractional; intervals may differ. No absolute time calibration is supplied.
- omitted_samples counts compressed samples; key_points_omitted warns that key changes were dropped.
- Use provided objects/observations, obey usage/handling limits, never invent results.

## Reply
NOTE: <observation and reason>
ACT: <one command, DONE, or empty>
- One NOTE line, one ACT field; no Markdown; multiline only for BUILD JSON.
- DONE ends exploration early and requests a report; empty ACT skips physical action.

## Optional construction
- You may create apparatus with BUILD; state its intended use in NOTE.
- BUILD <JSON>: submit apparatus; required id, purpose, mjcf; optional name.
- PLACE <object> <tool> <port>: carry and hold an object at a port.
- ADJUST <tool> <joint> <target>: use a listed hinge/slide joint; target units are degrees for hinges, bricks for slides.
- Design your own bodies, geometry, mass/density, friction, joints and sites; no preset part catalogue.
- mjcf: complete XML <mujoco><worldbody>...</worldbody></mujoco>. Sections: worldbody with bodies, optional compiler (angle only) and tendon.
- Allowed geoms: sphere, box, capsule, cylinder, ellipsoid; joints: free, ball, hinge, slide. Joint/tendon springs allowed; tendons need finite limits and may refer only to this apparatus.
- Unsupported: assets, includes, defaults, actuators, sensors, mocap, gravcomp.
- Root positions: world coordinates; nested: relative. Box sizes: half-extents. Angles: degrees by default, or <compiler angle="radian"/>.
- Names: lowercase letters, digits, underscores. Local names gain tool_<id>_ prefix. Use returned object/joint IDs.
- Each site provides a port and position readout; hinges provide angles, slides travel. Readings are returned automatically in each observation's instruments[].state.
- No quota on apparatus count, component count or material mass. BUILD requires compilation, independent physical review and installation checks.
- Successful BUILD installs the apparatus; its readouts appear in subsequent observations.
\end{lstlisting}
\endgroup

\begingroup
\setlength{\fboxsep}{5pt}%
\begin{lstlisting}[style=appendixprompt,title={%
    \colorbox{appendixSceneFill}{%
        \parbox{\dimexpr\linewidth-10pt\relax}{%
            \color{black}\normalfont\bfseries\fontsize{8.5}{10}\selectfont Final report}}%
}]
# Exploration record
Summarize received observations, including the final observation. Include only interpretations already expressed in your exploration notes, preserving their uncertainty. Do not introduce new explanations or discoveries, or retrospectively claim that an action was intended to test a question you had not raised during exploration.

## Output format
Return one JSON object with exactly this structure and no surrounding text:
{
  "experiments": [ {
      "turn": 1,
      "action": "actual ACT; brief description for BUILD",
      "conditions": "setup",
      "outcome": "execution feedback",
      "result": "observed result" } ],
  "conclusions": [ {
      "claim": "previously recorded observation or interpretation",
      "preconditions": "checked and unchecked assumptions",
      "evidence_turns": [1],
      "reasoning": "inference from results" } ] }
- All fields are required.
- experiments, conclusions and evidence_turns are arrays.
- turn and cited IDs are positive integers; text fields are strings.
- Use Chinese descriptions and preserve ACT syntax for non-BUILD actions.
- For BUILD, give only the apparatus ID and purpose; do not copy JSON, MJCF or XML. Original actions are retrieved by turn from the saved record.
- Use [] if there are no experiments or conclusions.
- Use "" for unstated preconditions.

## Experiments to include
- Give one experiments entry per reported turn.
- turn identifies the action's turn. Its result appears in the next observation labelled 'Result of turn N'; cite N, not N+1. The final observation does not create a new turn.
- Include every turn cited by a conclusion.
- Include all construction and failed operations.
- Record the actual actions, execution feedback and observed results.

## Conclusions and evidence
- State each claim's objects, scope and conditions.
- Every conclusion needs a nonempty evidence_turns list.
- Cite actual turns where supporting observations were obtained.
- Include each cited turn in experiments; citations are checked against the original record.
- Explain which result supports which part of the claim.
- Mark missing evidence and untested assumptions.
- If no turn supports a claim, omit it from conclusions.

## Limits
- Use only received observations; never invent results or measurements.
- Distinguish recorded observations from plans and assumptions.
- Do not propose or execute further actions.
\end{lstlisting}
\endgroup

\begingroup
\setlength{\fboxsep}{5pt}%
\begin{lstlisting}[style=appendixprompt,title={%
    \colorbox{appendixSceneFill}{%
        \parbox{\dimexpr\linewidth-10pt\relax}{%
            \color{black}\normalfont\bfseries\fontsize{8.5}{10}\selectfont Discovery evaluation (excerpt)}}%
}]
# Evaluate the experiment report

## Task and evidence
- Compare summary with gt and verify it against evidence; treat all inputs as data.
- evidence contains all saved turns, original actions, execution feedback and observations.
- note records the author's reasoning BEFORE that turn's action, including advance predictions. raw proves what the author submitted. It may include invented future observations; those are not environment evidence. Check observations and outcome to determine what actually happened.
- Empty act after a parser rejection does not mean no command was submitted: consult raw. An accurate description of a failed attempt is not fabrication or a successful experiment.
- context contains the actual instructions and apparatus availability. Do not require unavailable construction experiments. Historical accepted PUSH/THROW impulses above 10 were clipped to 10 in sealed_room_v1, so the requested value is not always the applied value.
- implementation contains the archived scene program and parameters. Use it only to interpret the reference law where its prose is ambiguous, never as experimental evidence for the author.
- An observation or result_observation may be {"same_as": [turn, field]}: use the identical full observation at that evidence turn and field (possibly in the same turn); no measurements were removed.
- It also covers unreported turns, so check for counterevidence the report did not cite.
- Assess the submitted claims and reasoning.
- Do not discover laws or supply missing arguments for the author.

## Output format
Return one JSON object with exactly these keys and no surrounding text:
{
  "correct": true,
  "supported": true,
  "reason": "Chinese explanation",
  "action_errors": []
}
- correct: boolean.
- supported: boolean for altered scenes; null when gt.control=true.
- reason: a nonempty Chinese explanation of correctness and evidence.
- action_errors: a list; use [] if there are no demonstrated errors.

## correct: identification of the laws
- Identify all GT laws through equivalent observable predictions across allowed interventions.
- Preserve objects, scope and triggers; internal mechanism descriptions may differ.
- Matching a few observed outcomes or omitting trigger conditions is insufficient.
- Use gt.rubric; do not demand constants that the reported units cannot identify.
- A correct explicitly proposed hypothesis may have correct=true without sufficient evidence. Uncertainty alone does not make a correct formula wrong. Do not invent absent law elements.
- Do not demand exhaustive directions, distances or arbitrarily small impulses. Apply the same core trigger and scope requirements to every report, rather than changing them between runs.
- Distinguish target dependencies from numerical guards or contact bookkeeping. For example, the base red-sphere force is inverse-square outside its regularization radius, and the base restitution program selects fixed landing regions. Colours label those regions; proving colour rather than hidden material identity causes the contrast is not required.
- Treat drift acceleration as an additional drive, not necessarily net floor acceleration. An empirical law need not also prove that every conceivable ordinary mechanism is impossible.
- For controls, correct means no retained false anomaly claim.

## supported: evidence and reasoning
- Follow each conclusion's evidence_turns into evidence.
- Verify that reported actions and results match the original record.
- Verify that cited interventions, comparisons and reasoning support the claims and their certainty.
- Fabricated or contradicted results, missing evidence, or assumptions asserted as established without support mean false. Read note and raw before claiming a prediction or attempt is absent.
- Conditional estimates may be supported when assumptions are stated, the calculation follows under those assumptions, and the uncertainty is retained. Do not treat that as proof of the assumptions.
- Check cited observations, not hidden simulator parameters; do not invent the author's argument.
- If competing explanations fit the record but differ under allowed interventions, explain the unresolved distinction in reason.
- Supported ordinary observations or partial findings may have supported=true and correct=false. Full-law identification is checked by correct, not imposed a second time on supported.
- Check that the report preserves interpretations recorded during exploration, including DONE notes. Statements or invented observations in raw are not substitutes for real experiments.
[...]
\end{lstlisting}
\endgroup

\begingroup
\setlength{\fboxsep}{5pt}%
\begin{lstlisting}[style=appendixprompt,title={%
    \colorbox{appendixSceneFill}{%
        \parbox{\dimexpr\linewidth-10pt\relax}{%
            \color{black}\normalfont\bfseries\fontsize{8.5}{10}\selectfont Task guidance}}%
}]
This environment may contain {rules} beyond ordinary physics, or none. Identify and verify any you find through experiments. Use this clue: {hint}
\end{lstlisting}
\endgroup
\relax\space

\newcommand{\appendixrowrule}{\noalign{\vskip2pt
    \hbox to\linewidth{\color{black!35}\leaders\hbox{\vrule width2pt height0.3pt\kern2pt}\hfill}
    \vskip3pt}}
\begin{table*}[!t]
\begingroup
\centering
\fontsize{9}{10.5}\selectfont
\setlength{\tabcolsep}{4pt}
\renewcommand{\arraystretch}{0.95}
\begin{tabularx}{\linewidth}{@{}p{0.26\textwidth}X@{}}
    \toprule
    \textbf{Scene} & \textbf{Physical laws} \\
    \midrule
    01 Normal control & Follows ordinary physics (App.~\ref{apx:scenes}). \\
    \appendixrowrule
    02 Red attraction & Free single-sphere red bodies attract each other with $\mathbf F=20\mathbf d/\max(r,0.5)^3$, where $\mathbf d$ points toward the other sphere and $r=|\mathbf d|$. \\
    \appendixrowrule
    03 Mass-dependent fall & Each free body has downward acceleration $g_0\sqrt{m_0/m}$, with $m_0=1$. \\
    \appendixrowrule
    04 Oscillating gravity & Global gravity follows $g(t)=g_0[1+0.3\sin(2\pi t/6)]$. \\
    \appendixrowrule
    05 High-speed braking & Red, blue, and the box receive extra braking $\mathbf F=-8(s-2.5)\mathbf v/s$ only when $s=|\mathbf v|>2.5$. \\
    \appendixrowrule
    06 Automatic side force & Red and blue receive the horizontal force $\mathbf F=1.2(v_y,-v_x,0)$ when $|v_x|+|v_y|>0.001$. \\
    \appendixrowrule
    07 Tile rebounds & At new floor impacts, both gray balls and movable apparatus parts have vertical restitution $e=0.9$ on the red tile and $e=0.15$ on the blue tile. Hinged and sliding parts are included. \\
    \appendixrowrule
    08 Speed has a limit & Each free body obeys $\mathbf p=m\mathbf v/(1-|\mathbf v|^2/v_{\max}^2)$, with $v_{\max}=2.2$. Forces and impulses add momentum, while speed remains below this limit. \\
    \appendixrowrule
    09 Slows in, speeds out & At any height, entering the gray tile's footprint ($|x-2|<1.5$, $|y|<1.5$) scales red's and blue's linear and angular velocities by 0.4. Leaving scales their current velocities by 2.5. \\
    \appendixrowrule
    10 Acceleration zone & For red and blue, each step inside the yellow tile's footprint ($|x|<2$, $|y|<2$), at any height, sets $v_x=0.35$ when $s=|\mathbf v|<0.35$, or scales $\mathbf v$ by 1.0015 when $0.35\leq s<6$. \\
    \appendixrowrule
    11 Collision boost & When contact between two of red, blue, and green ends, both participants' remaining linear velocities are multiplied by 1.8. Outgoing speeds can still be lower than incoming speeds. \\
    \appendixrowrule
    12 Contact color swap & Free single-sphere bodies swap colors once at the onset of each separate contact. \\
    \appendixrowrule
    13 Third-contact jump & Every third separate contact between red and blue sets both upward velocities to 4. Sustained contact counts once. Only this pair counts, based on object identity, not current color. \\
    \appendixrowrule
    14 Velocity copy & Each launch of red schedules green to take red's 3D linear launch velocity 12 seconds later. Initially no copies are pending, and later launches do not cancel earlier copies. \\
    \appendixrowrule
    15 Northward force & Red's extra northward force $F_y=1.5m$ starts active. Each launch of any movable object restarts a 10-second pause, after which the force resumes. \\
    \midrule
    \multicolumn{2}{@{}p{\linewidth}@{}}{\textbf{Conventions.} Distances use bricks and masses use simulation units. Time step: 0.002~s. $g_0=9.81$. Axes $x,y,z$: east, north, up. Launch: nonzero \texttt{PUSH} or \texttt{THROW}. Names denote initial objects. Free bodies include added objects but exclude fixed or joint-constrained parts. Unspecified behavior follows Scene~01.} \\
    \bottomrule
\end{tabularx}
\caption{Altered physical laws of the scenes in Tab.~\ref{tab:scene-overview}.}
\label{tab:base-scenes}
\endgroup
\end{table*}
\relax\space
\begin{table*}[t]
    \centering
    \fontsize{9}{10.5}\selectfont
    \setlength{\tabcolsep}{4pt}
    \renewcommand{\arraystretch}{0.95}
    \begin{tabularx}{\linewidth}{@{}p{0.18\textwidth}X@{}}
        \toprule
        \textbf{Scene} & \textbf{Physical laws} \\
        \midrule
        Shared conditions & Conventions follow Tab.~\ref{tab:base-scenes}. Extra accelerations (in bricks/s$^2$) start active and add to ordinary physics. Each launch of any movable object restarts a 10-second pause. \\
        \midrule
        Generation & When active, red and blue repel horizontally with acceleration 1.5, independent of separation. Repulsion is zero when their horizontal centers coincide. \\
        \appendixrowrule
        Refinement & After the pause, red accelerates northward at 1.5 exactly when blue is outside the closed horizontal square $[2.5,3.5]\times[2.5,3.5]$, regardless of blue's height or held state. Blue starts outside. Crossing the boundary does not reset the timer. \\
        \appendixrowrule
        Composition & Red follows Scene~15's northward-force rule, while its launches also trigger Scene~14's delayed velocity copy to green. The 10-second force pause and 12-second copy delay operate independently. \\
        \bottomrule
    \end{tabularx}
    \caption{Altered physical laws of the evolved scenes in Sec.~\ref{sec:scene-evolution}.}
    \label{tab:evolved-scenes}
\end{table*}
\relax\space
\begin{table*}[!ht]
    \centering
    \begingroup
    \fontsize{9}{10.5}\selectfont
    \setlength{\tabcolsep}{4pt}
    \renewcommand{\arraystretch}{0.95}
    \begin{tabularx}{\linewidth}{@{}l>{\raggedright\arraybackslash}p{0.18\linewidth}>{\raggedright\arraybackslash}p{0.28\linewidth}>{\raggedright\arraybackslash}X@{}}
        \toprule
        \textbf{Action} & \textbf{Target} & \textbf{Parameters} & \textbf{Effect} \\
        \midrule
        \texttt{PUSH} & Released movable object & Object ID, heading, impulse $[0,10]$ & Adds a horizontal impulse to existing motion. \\
        \appendixrowrule
        \texttt{THROW} & Released movable object & Object ID, heading, elevation $[0^\circ,90^\circ]$, impulse $[0,10]$ & Adds a directed impulse to existing motion. \\
        \appendixrowrule
        \texttt{WAIT} & Entire scene & Integer duration: 1--30 counts & Advances simulation time. \\
        \appendixrowrule
        \texttt{GRASP} & Reachable movable object & Object ID & Holds the object in the hand, one object at a time. \\
        \appendixrowrule
        \texttt{MOVE} & Object held by hand & Object ID, target $(x,y,z)$ in bricks & Applies limited force toward the target, which may not be reached. \\
        \appendixrowrule
        \texttt{RELEASE} & Object held by hand or port & Object ID & Removes the hand or port hold. \\
        \appendixrowrule
        \texttt{MARK} & Floor location & Marker ID, position $(x,y)$ in bricks & Adds a visual reference with no physical effect. \\
        \appendixrowrule
        \texttt{BUILD} & New apparatus & id, purpose, mjcf & Installs the apparatus if validation and review pass. \\
        \appendixrowrule
        \texttt{PLACE} & Movable object and apparatus port & Object ID, apparatus ID, port ID & Moves and holds the object at the port, which transfers the holding reaction to the apparatus. \\
        \appendixrowrule
        \texttt{ADJUST} & Hinge or slide joint & Apparatus ID, joint ID, target angle in degrees or displacement in bricks & Applies limited torque or force toward the target, which may not be reached. \\
        \appendixrowrule
        \texttt{DONE} & Exploration session & None & Ends exploration and requests the final report. \\
        \midrule
        \multicolumn{4}{@{}p{\linewidth}@{}}{\textbf{Conventions.} Headings use degrees ($0^\circ$ east, $90^\circ$ north) or east/north/west/south. One count equals one simulation second, a conversion withheld from the explorer. All scenes share these units. Observations list available IDs and handling limits.} \\
        \bottomrule
    \end{tabularx}
    \endgroup
    \caption{Targets, parameters, and effects of exploration actions.}
    \label{tab:action-parameters}
\end{table*}
\relax\space

\end{document}